\documentclass[10pt,twocolumn,letterpaper]{article}

\usepackage{cvpr}              

\usepackage{graphicx}
\usepackage{amsmath}
\usepackage{amssymb}
\usepackage{subcaption}
\usepackage{enumerate}
\usepackage{ulem}
\usepackage{makecell}

\usepackage{amsmath}
\usepackage{booktabs} 
\usepackage{graphicx}
\usepackage{color}
\usepackage{float}
\usepackage{array}
\usepackage{multirow}
\usepackage{amssymb}
\usepackage{graphbox}

\usepackage{times}
\usepackage{epsfig}
\usepackage{graphicx}
\usepackage{amsmath}
\usepackage{amssymb}

\usepackage{url}            
\usepackage{booktabs}       
\usepackage{amsfonts}       
\usepackage{comment}
\usepackage{color}
\usepackage{multirow}
\usepackage{array}
\usepackage{graphbox}
\usepackage{booktabs} 
\usepackage{wrapfig}
\usepackage{enumitem}
\usepackage{bbding}

\usepackage{times,graphicx,amsmath,amssymb,booktabs,tabulary,multirow,overpic,xcolor}
\usepackage[pagebackref=true,breaklinks=true,letterpaper=true,colorlinks,bookmarks=false]{hyperref}

\renewcommand{\vec}[1]{\boldsymbol{#1}}

\newcommand{\bignum}[1]{\uppercase\expandafter{\romannumeral#1}}
\usepackage[british,english,american]{babel}

\newcommand{\app}{\raise.17ex\hbox{$\scriptstyle\sim$}}

\newcolumntype{x}[1]{>{\centering\arraybackslash}p{#1pt}}

\newlength\savewidth

\makeatletter\renewcommand\paragraph{\@startsection{paragraph}{4}{\z@}
	{.5em \@plus1ex \@minus.2ex}{-.5em}{\normalfont\normalsize\bfseries}}\makeatother

\usepackage[capitalize]{cleveref}
\crefname{section}{Sec.}{Secs.}
\Crefname{section}{Section}{Sections}
\Crefname{table}{Table}{Tables}
\crefname{table}{Tab.}{Tabs.}

\def\cvprPaperID{1234} 
\def\confName{CVPR}
\def\confYear{2022}
\begin{document}
	
	\title{Generalized Few-shot Semantic Segmentation}
	
	\author{
		Zhuotao Tian$^{1}$
		\hspace{0.4cm}Xin Lai$^{1}$
		\hspace{0.4cm}Li Jiang$^{2}$
		\hspace{0.4cm}Shu Liu$^{3}$
		\hspace{0.4cm}Michelle Shu$^{4}$
		\hspace{0.4cm}Hengshuang Zhao$^{5,6}$
		\hspace{0.4cm}Jiaya Jia$^{1,3}$
		\vspace{0.2cm}\\
		$^{1}$CUHK~~~
		$^{2}$MPI Informatics~~~
		$^{3}$SmartMore~~~
		$^{4}$Cornell University~~~
		$^{5}$HKU~~~
		$^{6}$MIT
	}
	
\maketitle

\begin{abstract}
Training semantic segmentation models requires a large amount of finely annotated data, making it hard to quickly adapt to novel classes not satisfying this condition. Few-Shot Segmentation (FS-Seg) tackles this problem with many constraints. In this paper, we introduce a new benchmark, called Generalized Few-Shot Semantic Segmentation (GFS-Seg), to analyze the generalization ability of simultaneously segmenting the novel categories with very few examples and the base categories with sufficient examples. It is the first study showing that previous representative state-of-the-art FS-Seg methods fall short in GFS-Seg and the performance discrepancy mainly comes from the constrained setting of FS-Seg. To make GFS-Seg tractable, we set up a GFS-Seg baseline that achieves decent performance without structural change on the original model. Then, since context is essential for semantic segmentation, we propose the Context-Aware Prototype Learning (CAPL) that significantly improves performance by 1) leveraging the co-occurrence prior knowledge from support samples, and 2) dynamically enriching contextual information to the classifier, conditioned on the content of each query image. Both two contributions are experimentally shown to have substantial practical merit. Extensive experiments on Pascal-VOC and COCO manifest the effectiveness of CAPL, and CAPL generalizes well to FS-Seg by achieving competitive performance. Code is available at \url{https://github.com/dvlab-research/GFS-Seg}. 
\end{abstract}

\maketitle

\section{Introduction}

\begin{figure}
	\centering
	\begin{minipage}   {1\linewidth}
		\centering
		\includegraphics [width=1\linewidth] 
		{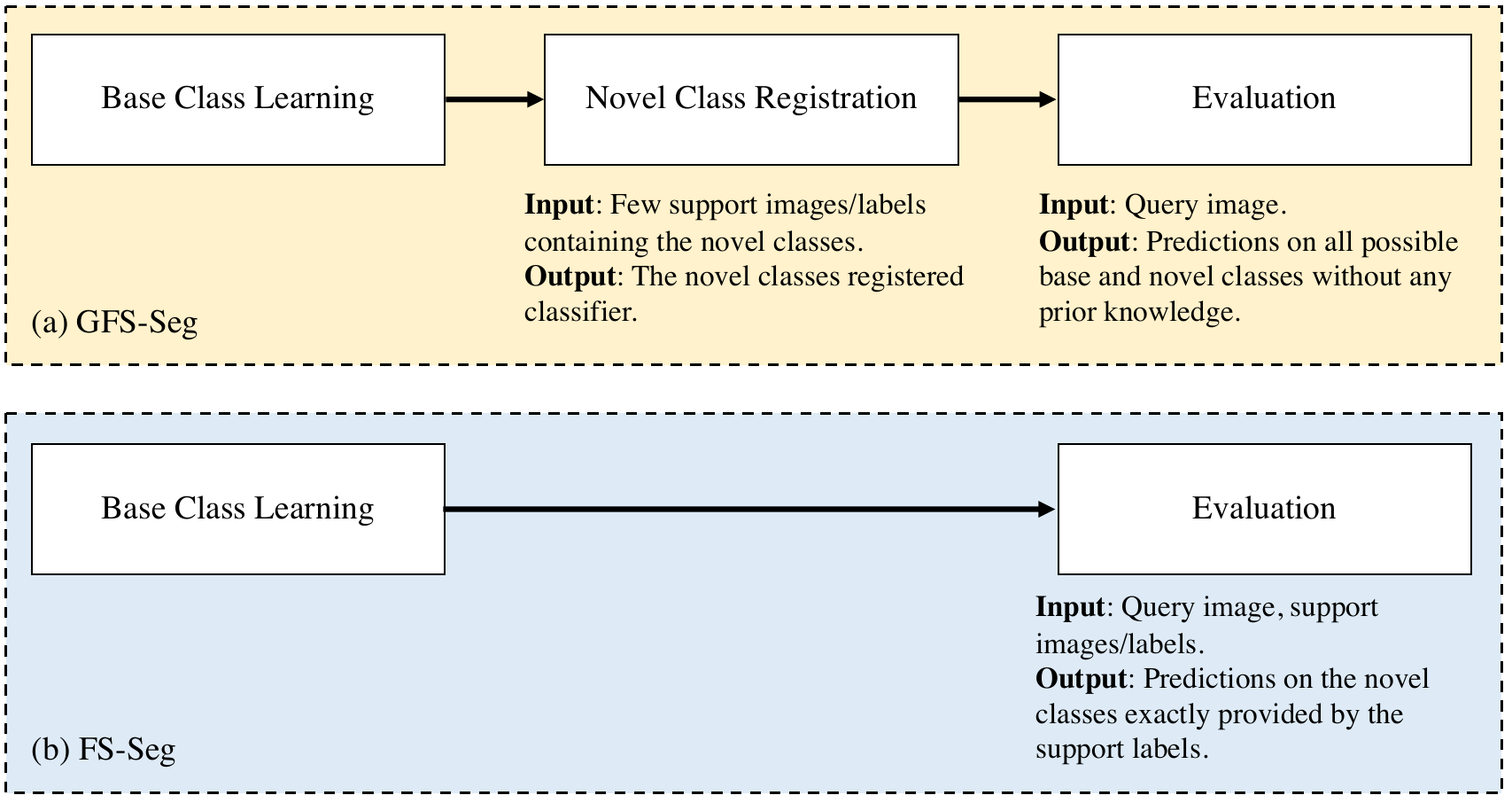}
	\end{minipage}    
	\caption{Pipeline illustrations of GFS-Seg and FS-Seg. (a) GFS-Seg has an additional \textit{novel class registration phase} that registers the novel information to the new classifier, therefore, in the last \textit{evaluation phase}, GFS-Seg methods are able to make predictions on all possible base and novel classes as testing normal segmentation models without forwarding additional support samples that provide the prior knowledge of target classes. Contrarily, (b) FS-Seg models in the \textit{evaluation phase} require support images/labels to provide the information of target classes exactly contained in each query image.}
	\label{fig:concept_compare}
	\vspace{-0.4cm}
\end{figure}

The development of deep learning has provided significant performance gain to semantic segmentation tasks. Representative semantic segmentation methods \cite{pspnet,deeplab} have benefited a wide range of applications for robotics, automatic driving, medical imaging, etc. However, once these frameworks are trained, without sufficient fully-labeled data, they are unable to deal with unseen classes in new applications. Even if the required data of novel classes are ready, fine-tuning costs additional time and resources.

In order to quickly adapt to novel classes with only limited labeled data, Few-Shot Segmentation (denoted as FS-Seg) \cite{shaban} models are trained on well-labeled base classes and then tested on previously unseen novel classes. 
During training, FS-Seg divides data into the support and query sets. Samples of support set aim to provide FS-Seg models with target categorical information to identify target regions in query samples, with the purpose to mimic the situation where only a few labeled data of novel classes are available. After training, both support and query samples are sent to FS-Seg models to yield query predictions on previously unseen classes based on the support information.

\vspace{-0.15cm}
\paragraph{Limitations of FS-Seg. }
However, FS-Seg requires support samples to contain classes that exist in query samples. It may be overly strong in many situations to have this prior knowledge because providing support samples in the same classes requires cumbersome manual selection. Besides, FS-Seg only evaluates the novel classes, while test samples in normal semantic segmentation may also contain the base classes. Experiments show that exemplar FS-Seg models cannot well tackle the practical situation of evaluation on both base and novel classes due to these constraints.

\vspace{-0.1cm}
\paragraph{New benchmark and our solution. }
With these facts, we set up a new task, named Generalized Few-Shot Semantic Segmentation (GFS-Seg). As shown in Figure~\ref{fig:concept_compare}, a typical GFS-Seg method should have three phases: 1) \textit{Base class learning phase}, 2) \textit{novel class registration phase} with few support samples containing novel classes, and 3) \textit{evaluation phase} on both base and novel classes. The difference between GFS-Seg and FS-Seg is that, during evaluation, GFS-Seg does not require forwarding support samples that contain the same target classes in the test (query) samples to make predictions, because GFS-Seg should have obtained the information of base and novel classes during the \textit{base class learning phase} and \textit{novel class registration phase} respectively. GFS-Seg can perform well on novel classes without sacrificing the accuracy of base classes when making predictions on them simultaneously without knowing what classes are contained in the query images in advance, achieving the essential step towards practical use of semantic segmentation in more challenging situations.

Inspired by \cite{dynamic_noforget, imprint_cvpr18}, we design a baseline for GFS-Seg with decent performance. Considering the contextual relation is essential for semantic segmentation, we further propose the Context-Aware Prototype Learning (CAPL) that provides significant performance gain to the baseline by updating the weights of base prototypes with adapted features. CAPL not only exploits essential co-occurrence information from support samples, but it also adapts the model to various contexts of query images. The baseline method and the proposed CAPL can be applied to normal semantic segmentation models, \eg, FCN~\cite{fcn}, PSPNet~\cite{pspnet} and DeepLab~\cite{deeplab}. Also,  CAPL demonstrates its effectiveness in the setting of FS-Seg by improving the baseline by a large margin, reaching state-of-the-art performance. Our overall contributions are as follows.  \\
\vspace{-0.4cm}
\begin{itemize}[leftmargin=20pt]
	\item We extend the classic Few-Shot Segmentation (FS-Seg) and propose a more practical setting -- Generalized Few-Shot Semantic Segmentation (GFS-Seg). 
	\item  Based on experimental results, we give an analysis on the existing performance gap between FS-Seg and GFS-Seg that even the recent popular FS-Seg models cannot well tackle.
	\item We propose the Context-Aware Prototype Learning (CAPL) that 
	brings significant performance gains to the baseline models 
	in both settings of GFS-Seg and FS-Seg, and it is general to be applied to various normal semantic segmentation models without specific structural constraints.
\end{itemize}

\section{Related Work}
\paragraph{Semantic segmentation. }
Semantic segmentation is a fundamental while challenging topic that asks models to accurately predict the label for each pixel. FCN~\cite{fcn} is the first framework designed for semantic segmentation by replacing the last fully-connected layer in a classification network with convolution layers. To get per-pixel predictions, some encoder-decoder style approaches~\cite{deconvnet,segnet,unet} are adopted to help refine the outputs step by step. The receptive field is vital for semantic segmentation thus dilated convolution~\cite{deeplab,dilation} is introduced to enlarge the receptive field. Context information plays an important role for semantic segmentation and some context modeling architectures are introduced like global pooling~\cite{parsenet} and pyramid pooling~\cite{deeplab,pspnet,icnet,denseaspp}. Meanwhile, attention models~\cite{psanet,encnet,danet,hrnet_pami,hrnet_cvpr,ocr,ccnet} are also shown to be effective for capturing the long-range relationships inside scenes. Despite the success of these powerful segmentation frameworks, they cannot be easily adapted to unseen classes without fine-tuning on sufficient annotated data.

\paragraph{Few-shot learning. } 
Few-shot learning aims at making prediction on novel classes with only a few labeled examples. Popular solutions include meta-learning based methods \cite{memory_match,maml,leo} and metric-learning ones \cite{matchingnet,relationnet,prototype_cls,deepemd}. 
In addition, data augmentation helps models achieve better performance by combating overfitting. Therefore, synthesizing new training samples or features based on a few labeled data is also a feasible solution for tackling the few-shot problem \cite{hallucinate_saliency,hallucinating,imaginary}.  
Generalized few-shot learning was proposed in \cite{hallucinating} where the query images can be from either base or novel categories. 
Though the combination of a supervised model (for base classes) and a prototype-based approach (for novel classes) has already been explored in work on low-shot visual recognition~\cite{dynamic_noforget,imprint_cvpr18}, the dense pixel labeling in semantic segmentation is different from the image-level classification that does not contain contextual information for each target. 

\paragraph{Few-shot segmentation. } 
Few-Shot Segmentation (FS-Seg) places semantic segmentation in the few-shot scenario \cite{adaptivemaskweightimprinting,guide,AttantionMCG,
	classrep,FGN,selftuning,pr_logo,mmm,crnet,differentiable,simpropnet,democratic,fss1000,objectness,Zhang_2019_ICCV,SG-One,amp,co-FCN,ppnet,protomix,weighingboosting,tnnls}, where dense pixel labeling is performed on new classes with only a few support samples. 
OSLSM \cite{shaban} first introduces this setting in segmentation and provides a solution by yielding weights of the final classifier for each query-support pair during evaluation. The idea of the prototype is used in PL \cite{prototype_seg} where predictions are based on the cosine similarity between pixels and the prototypes. Also, prototype alignment regularization is introduced in PANet \cite{panet}. Predictions can be also generated by convolutions. PFENet \cite{pfenet} uses the prior knowledge from a pre-trained backbone to help find the region of the interest, and the spatial inconsistency between the query and support samples is alleviated by Feature Enrichment Module (FEM).

Though FS-Seg models perform well on identifying novel classes given the corresponding support samples, as shown in Section \ref{sec:experiments}, without the prior knowledge of the target classes contained in query images, even the state-of-the-art FS-Seg model cannot well tackle the practical setting that contains both base and novel classes.

\section{Towards Generalized Few-shot Semantic Segmentation}
\label{sec:task_description}

\begin{figure*}[t]
	\centering
	\begin{tabular}{@{\hspace{0.0mm}}c@{\hspace{3.0mm}}c@{\hspace{0.0mm}}}
		\includegraphics[align=c, width=0.49\linewidth]{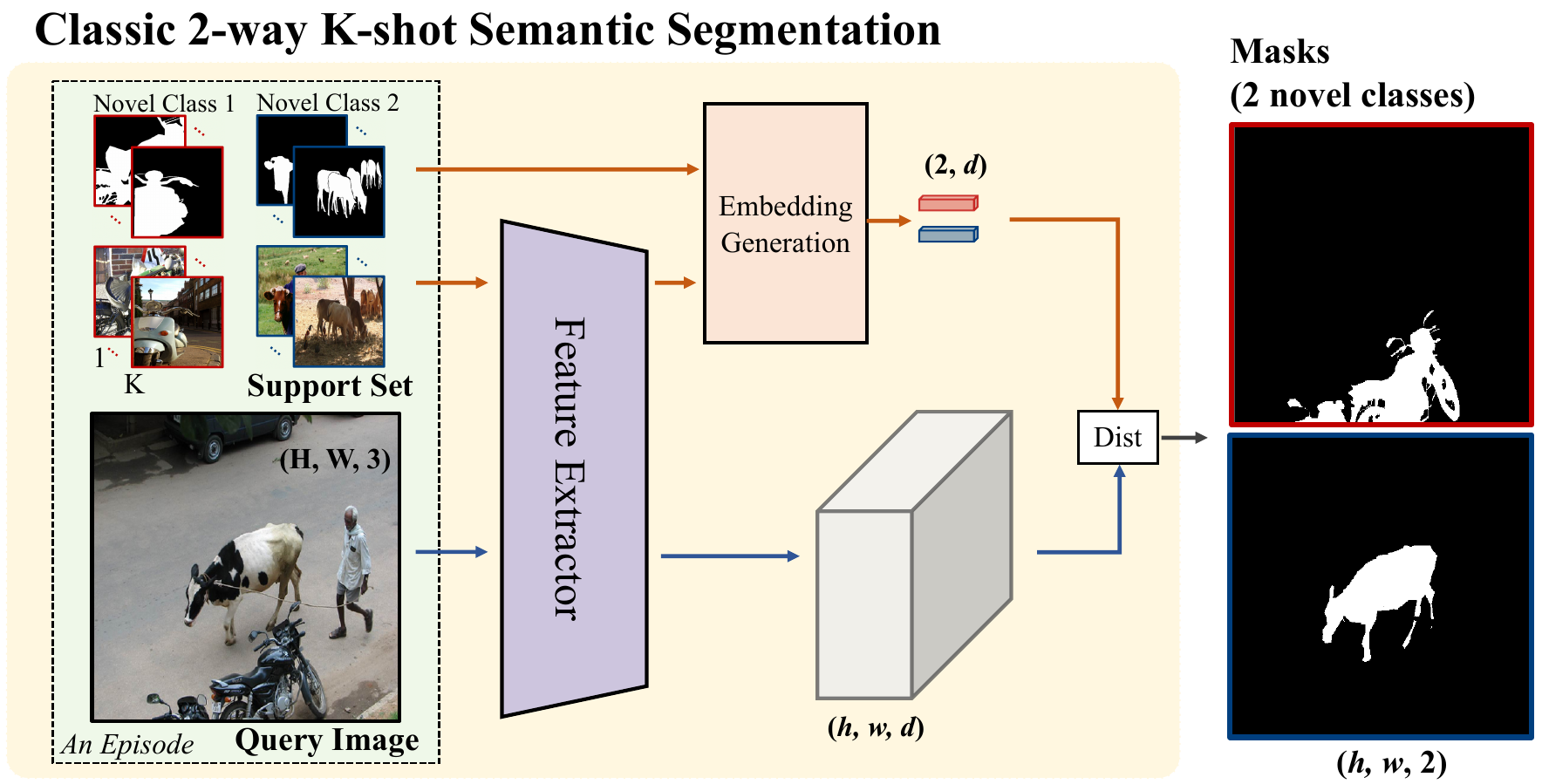}&
		\includegraphics[align=c, width=0.49\linewidth]{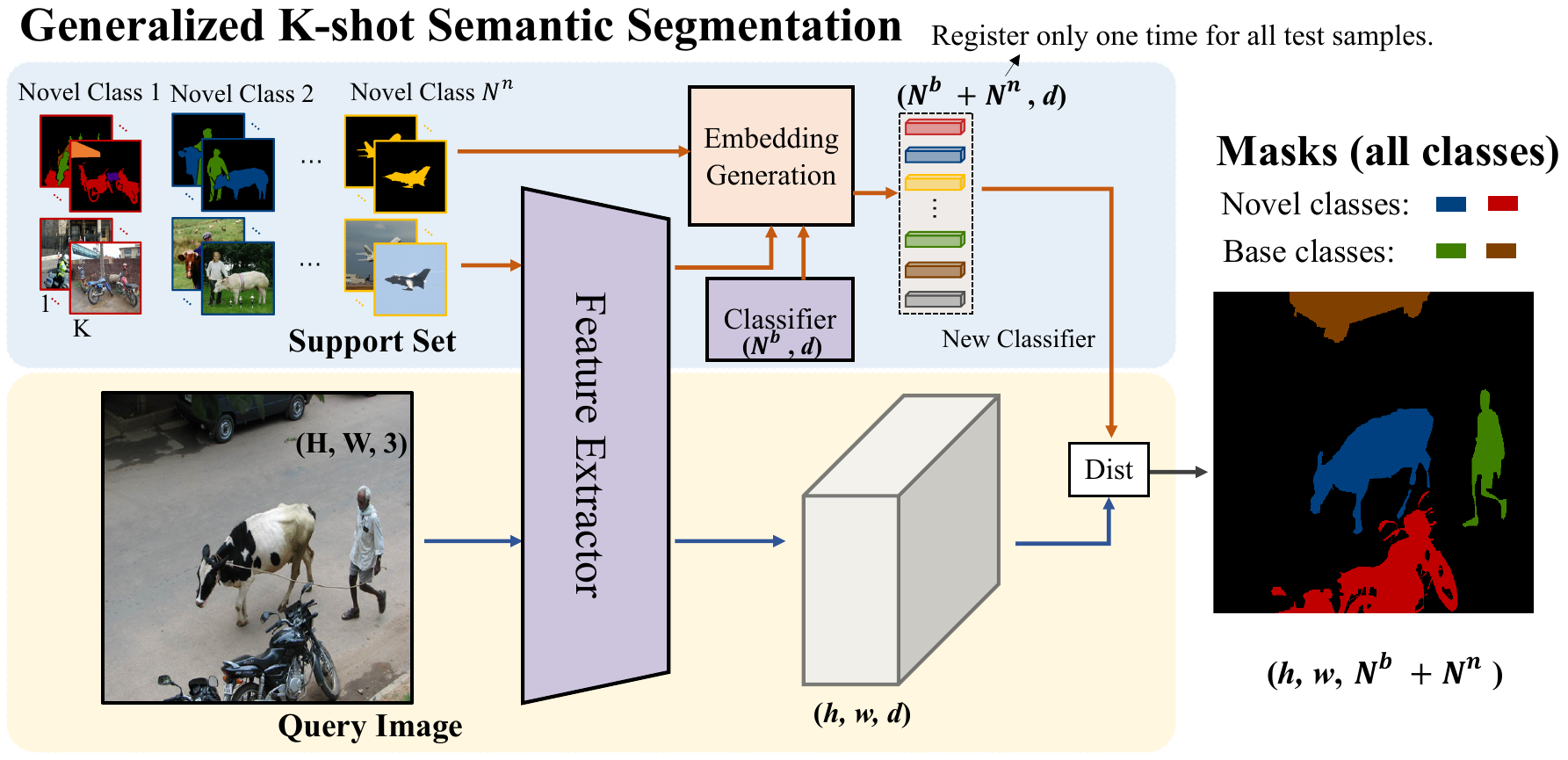} \\
		(a)&(b)
	\end{tabular}
	\caption{Illustrations of (a) classic Few-Shot Segmentation (FS-Seg) and (b) Generalized Few-Shot Semantic Segmentation (GFS-Seg). `Dist' can be any method measuring the distance/similarity between each feature and prototype and making predictions based on that distance/similarity. FS-Seg models only predict novel classes provided by the support set, while GFS-Seg models make predictions on base and novel classes simultaneously without being affected by redundant classes. Also, during evaluation, GFS-Seg models do not require the prior knowledge of what target classes exist in the query images by registering the novel classes to form a new classifier one time for all test images (blue area in (b)  represents the novel class registration phase).}
	\label{fig:setting_compare}
	\vspace{-0.3cm}
\end{figure*}

\paragraph{Revisit the classic setting. } 
In classic Few-Shot Segmentation (FS-Seg)~\cite{shaban},  data is split into two sets for support $\vec{S}$ and query $\vec{Q}$. An FS-Seg model needs to make predictions on $\vec{Q}$ based on the class information provided by $\vec{S}$, and it is trained on base classes $\vec{C}^{b}$ in the \textit{base class learning phase} and tested on previously unseen novel classes $\vec{C}^{n} (\vec{C}^{b} \cap \vec{C}^{n} = \varnothing)$ in the \textit{evaluation phase}. To help adapt to novel classes, the episodic paradigm was proposed in \cite{matchingnet} to train and evaluate few-shot models. Each episode is formed by a support set $\vec{S}$ and a query set $\vec{Q}$ of the same class $c$. In a $K$-shot task, the support set $\vec{S}$ contains $K$ samples $\vec{S}=\{\vec{S}_1, \vec{S}_2, ..., \vec{S}_K\}$ of class $c$. Each support sample $\vec{S}_i$ is a pair of $\{\vec{s}_i, \vec{m}_i\}$ where $\vec{s}_i$ and $\vec{m}_i$ are the support image and mask of class $c$. 

For the query set, $\vec{Q}=\{\vec{q}, \vec{y}\}$ where $\vec{q}$ is the input query image and $\vec{y}$ is the ground truth mask of class $c$. The input data batch used for model training is the query-support pair $\{\vec{q}; \vec{S}\} = \{\vec{q}; \vec{s}_1, \vec{m}_1, ..., \vec{s}_k, \vec{m}_k\}$. The ground truth mask $\vec{y}$ of $\vec{q}$ is not accessible because it is to evaluate the prediction of the query image in each episode. More details for the episodic training paradigm in FS-Seg are included in \cite{shaban}.

In summary, there are two key criteria in the classic Few-Shot Segmentation task. (1) Samples of testing classes $\vec{C}^{n}$ are not seen by the model during training. (2) The model requires its support samples to contain target classes existing in query samples to make the corresponding predictions. 

\paragraph{The proposed generalized setting. }
Criterion (1) of classic few-shot segmentation evaluates model generalization ability to new classes with only a few provided samples. Criterion (2) makes this setting not practical in many cases since users may not know exactly how many and what classes are contained in each test image. So it is hard to feed the support samples that contain the same classes as the query sample to the model. 

Besides, even if users have already known that there are $N$ classes contained in the test image,
FS-Seg models \cite{pfenet,hsnet,repri,canet,scl,asgnet,Zhang_2019_ICCV} may need to process $NK$ additional manually selected support images/labels to make predictions over all possible classes for the test image. 
This is insufficient in real applications, 
where models are supposed to directly output predictions of all possible classes in the test image without processing $NK$ additional support images/labels that provide the prior knowledge of what classes are contained in the query image.

In GFS-Seg, base classes $\vec{C}^{b}$ have sufficient labeled training data and each novel class $\vec{C}^{n}_i \in \vec{C}^{n}$ only has limited $K$ labeled samples (\eg, $K=1, 5, 10$). Similar to FS-Seg, models in GFS-Seg are first trained on base classes $\vec{C}^{b}$ to learn good representations, which is the first \textit{base class learning phase} in GFS-Seg. Then, when the first phase is accomplished, the model acquires the information of $N$ novel classes from the limited $NK$ support samples and forms a new classifier (\ie, \textit{novel class registration phase}). In the last \textit{evaluation phase}, GFS-Seg models are evaluated on images of the test set to predict labels from both base and novel classes $\vec{C}^{b} \cup \vec{C}^{n}$, rather than only evaluating novel classes $\vec{C}^{n}$ as FS-Seg. 
Query images in GFS-Seg may contain either the novel classes, base classes, or both, and there is no prior knowledge of what classes are contained in the query images.
Therefore, the major evaluation metric of GFS-Seg is the total mIoU that is averaged over all classes rather than the novel mIoU used in FS-Seg. The episodic training paradigm might not be a good choice in GFS-Seg, since the input data during \textit{evaluation phase} is no longer the query-support pair $\{\vec{q};  \vec{S}\}$ used in FS-Seg, and instead is only the query image $I_Q$ as testing common semantic segmentation models.

To better distinguish between FS-Seg and GFS-Seg, we illustrate a 2-way $K$-shot task of FS-Seg and a case of GFS-Seg with the same query image in Figure \ref{fig:setting_compare}, where \textit{Cow} and \textit{Motorbike} are novel classes, and \textit{Person} and \textit{Car} are base classes. 
FS-Seg model (Figure \ref{fig:setting_compare} (a)) is limited to predicting binary segmentation masks only for the classes that are included in the support set. \textit{Person} on the right and \textit{Car} at the top are missing in the predictions because the support set does not provide information for these classes, even if the model has been trained on these base classes for sufficient steps.
In addition, if redundant novel classes that do not appear in the query image (\eg, \textit{Aeroplane}) are provided by support set of (a), they may adversely affect performance because FS-Seg has a prerequisite that the query images must contain the classes provided by support samples. 

As shown in Section \ref{sec:fs_seg_compare}, FS-Seg models only learn to predict the foreground masks for the given novel classes, so their performance is much degraded in our proposed generalized setting of GFS-Seg where all possible base and novel classes require predictions. Differently, GFS-Seg (Figure \ref{fig:setting_compare} (b)) identifies base and novel classes simultaneously 
without the prior knowledge of classes contained in query image,
and extra support classes (\eg, \textit{Aeroplane} at the top-left of Figure \ref{fig:setting_compare} (b)) should not affect the model much.

\section{Our Method}
\label{sec:method}

\subsection{Prototype Learning}
\paragraph{Prototype learning in FS-Seg. }
In Few-Shot Segmentation (FS-Seg) frameworks \cite{prototype_seg,panet} of a $N$-way $K$-shot FS-Seg task ($N$ novel classes and each novel class has $K$ support samples), all support samples $\vec{s}^i_j$ ($i\in\{1,2,...,N\}, j\in\{1,2,...,K\}$) are first processed by a feature extractor $\mathcal F$ and mask average pooling. They are then averaged over $K$ shots to form $N$ prototypes $\vec{p}^{i}$ ($i\in\{1,2,...,N\}$). 
\begin{equation}
	\label{eqn:extract_feat_seg}
	\footnotesize
	\vec{p}^{i} = \frac{1}{K} * \sum_{j=1}^{K} \frac{\sum_{h,w} [\vec{m}^i_j \circ  \mathcal F(\vec{s}^i_j)]_{h,w}}
	{\sum_{h,w} [\vec{m}^i_j]_{h,w}}, \quad i\in\{1,2,...,N\},
\end{equation}
where $\vec{m}^i_j \in \mathbb{R}^{h,w,1}$ is the class mask for class $c^i$ on $\mathcal F(\vec{s}^i_j)\in \mathbb{R}^{h,w,d}$. $\vec{s}^i_j$ represents the $j$-th support image of class $c^i$, and $\circ$ is the Hadamard Product. After acquiring $N$ prototypes, for query features $\mathcal F(\vec{q})$, the predictions are assigned by the class labels of the most similar prototypes. 

\paragraph{The baseline for GFS-Seg. }
FS-Seg only requires identifying targets from novel classes, but our generalized setting in GFS-Seg requires predictions on both base and novel classes. However, it is hard and also inefficient for the FS-Seg model to form prototypes for base classes via Eq. \eqref{eqn:extract_feat_seg} by forwarding all samples of base classes to the feature extractor, especially when the training set is large. 

The common semantic segmentation frameworks can be decomposed into two parts: feature extractor and  classifier. Feature extractor projects input image into $d$-dimensional latent space and then the classifier of size $N^{b} \times d$ makes predictions on $N^{b}$ base classes. In other words, the classifier of size $N^{b} \times d$ can be seen as $N^{b}$ base prototypes ($\vec{P}^{b} \in \mathbb{R}^{N^{b},d}$). Since forwarding all base samples to form the base prototypes is impractical, inspired by the low-shot learning methods \cite{dynamic_noforget,imprint_cvpr18}, we learn the classifier via back-propagation during training on base classes. 

Specifically, our baseline for GFS-Seg is trained on base classes as normal segmentation frameworks.
After training, the prototypes $\vec{P}^{n} \in \mathbb{R}^{N^{n}, d}$ for $N^{n}$ novel classes are formed as Eq. \eqref{eqn:extract_feat_seg} with $N^{n} \times K$ support samples. $\vec{P}^{n}$  and $\vec{P}^{b}$ are then concatenated to form $\vec{P}^{all} \in \mathbb{R}^{N^{b}+N^{n}, d}$, the new classifier, to simultaneously predict the base and novel classes.
Since the dot product used by classifier of common semantic segmentation frameworks produces different norm scales of $\mathcal F(\vec{s}^i_j)$ that negatively impact the average operation in Eq. \eqref{eqn:extract_feat_seg}, following \cite{panet}, we adopt cosine similarity as the distance metric $\phi$ to yield output $\vec{O}$ for pixels in query sample $\vec{q} \in \mathbb{R}^{h,w,3}$ as 
\begin{equation}
	\label{eqn:cosine_output}
	\footnotesize
	\vec{O}_{x,y} = \mathop{\arg\max}_{i}
	\frac
	{\exp(\alpha \phi(\mathcal F(\vec{q}_{x,y}), \vec{p}^i) )}
	{\sum_{\vec{p}^i \in \vec{P}^{all}} \exp(\alpha \phi (\mathcal F(\vec{q}_{x,y}), \vec{p}^i))},
\end{equation}
where $x \in \{1,...,h\},  y\in \{1,...,w\},   i \in \{1,...,N^b + N^n\}$, and $\alpha$ is set to 10 in all experiments.

\begin{figure*}
	\centering
	\centering
	\includegraphics [width=0.8\linewidth]
	{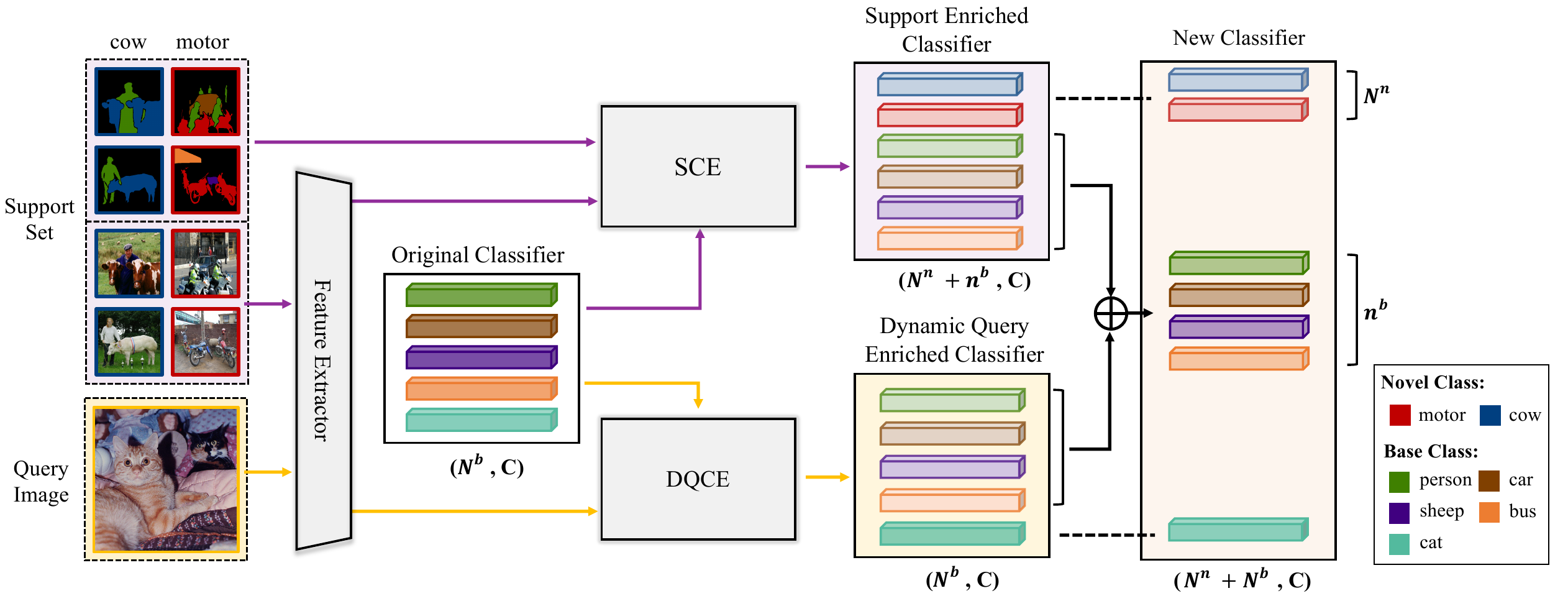}      
	\caption{Visual illustration of CAPL. The weights of $N^n$ novel classes (\eg \textit{motor} and \textit{cow}) are directly set by the averaged novel features. Also, the weights of $n^b$ base classes (\eg \textit{person}, \textit{car}, \textit{sheep} and \textit{bus}) that appear in support samples are enriched by SCE with the original weights. Besides, DQCE dynamically enriches the weights of $N^b$ base classes in the classifier with temporary contextual hints extracted from query samples. The new classifier takes the merits from both SCE and DQCE.}
	\label{fig:capl_example}
	\vspace{-0.5cm}
\end{figure*}

\subsection{Context-Aware Prototype Learning (CAPL)}
\paragraph{Motivation.} 
Prototype Learning (PL) is applicable to few-shot classification and FS-Seg, but it works inferiorly for GFS-Seg. In the setting of FS-Seg, target labels of query samples are only from novel classes. Thus there is no essential co-occurrence interaction between novel and base classes that can be utilized for further improvement. However, in GFS-Seg, there is no such limitation on the classes contained in each test image -- predictions on all possible base and novel classes are required.  

Contextual cue always plays an important role in semantic segmentation (\eg, PSPNet \cite{pspnet} and Deeplab \cite{deeplabv3+}), especially in the proposed setting (GFS-Seg).
For example, \textit{Dog} and \textit{People} are base classes. The learned base prototypes only capture the contextual relation between \textit{Dog} and \textit{People} during training. If \textit{Sofa} is a novel class and some instances of \textit{Sofa} in support samples appear with \textit{Dog} (\eg, a dog is lying on the sofa), 
merely mask-pooling in each support sample of \textit{Sofa} to form the novel prototype may result in the base prototype of  \textit{Dog} losing the contextual co-occurrence information with  \textit{Sofa}
and hence yield inferior results. Thus, for GFS-Seg, reasonable utilization of contextual information is the key to better performance, and our proposed method enables the model to meta-learn the behavior that the contextual information can be enriched by a simple adaptive fusion strategy without significantly altering the model structure. The contextual cues can be mined from both support and query samples. Therefore, we propose the Context-Aware Prototype Learning (CAPL) to tackle GFS-Seg by effectively enriching the classifier with contextual information. To facilitate understanding, the pipeline illustration of CAPL is shown in Figure \ref{fig:capl_example}.

\paragraph{Support Contextual Enrichment (SCE). }
 SCE mines support context in the \textit{novel class registration phase}. Let $N^b$ and $N^n$ denote the numbers of base classes $\vec{C}^{b}$ and novel classes $\vec{C}^{n}$ respectively. 
We use $n^b (n^b \leqslant N^b)$ to denote the number of base classes $c^{b,i} \in \vec{C}^{b} (i\in\{1,...,n^b\})$ contained in $N^n \times K$ support samples for $N^n$ unseen novel classes $c^{n,u} \in \vec{C}^{n} (u \in \{ 1,...,N^n \})$.
Before evaluation, the novel prototypes are formed as Eq. \eqref{eqn:extract_feat_seg}. The updated prototype $\vec{p}^{b, i}$ for base class $c^{b,i}$ is the weighted sum of the weights of original classifier $\vec{p}^{b, i}_{cls}$ and the new base prototypes $\vec{p}^{b, i}_{sup}$ generated from support features, written as
\begin{equation}
    \footnotesize
	\begin{aligned}
		\label{eqn:capl_base_new_proto}
		\vec{p}^{b, i}_{sup} = \frac
		{\sum_{u=1}^{N^{n}} \sum_{j=1}^{K} \sum_{h,w} [\vec{m}_j^{i,u} \circ  \mathcal F(\vec{s}^u_j)]_{h,w}}   
		{\sum_{u=1}^{N^{n}} \sum_{j=1}^{K} \sum_{h,w} [\vec{m}_j^{i,u}]_{h,w} }, 
		\\
		i\in\{1,...,n^b\}, u \in \{1,...,N^n\}
	\end{aligned}
\end{equation}
\begin{equation}
    \footnotesize
	\label{eqn:capl_base_proto}
	\vec{p}^{b, i} = \vec{\gamma}^i_{sup} * \vec{p}^{b, i}_{cls} + (1 - \vec{\gamma}^i_{sup}) * \vec{p}^{b, i}_{sup}
	, \quad i\in\{1,...,n^b\},
\end{equation}
where $\vec{m}_j^{i,u}$ represents the binary mask for base class $c^{b,i}$ on $\mathcal F(\vec{s}^u_j)$, and $\vec{s}^u_j$ is the $j$-th support sample of novel class $c^{n,u}$. The adaptive weight $\vec{\gamma}^i_{sup}$ balances the impacts of the old and new base prototypes. 

We note that $\vec{\gamma}^i_{sup}$ is data-dependent and its value is conditioned on each pair of old and new prototypes. 
For base class $c^{b,i}$, $\vec{\gamma}^i_{sup}$ is calculated as $\vec{\gamma}^i_{sup} = \mathcal G_{sup}(\vec{p}^{b, i}_{cls}, \vec{p}^{b, i}_{sup})$ where $\mathcal G_{sup}$ serves as a correlation estimator that produces the weighing factor $\vec{\gamma}^i_{sup}$.

\paragraph{Dynamic Query Contextual Enrichment (DQCE). }
The aforementioned SCE only takes place in the \textit{novel class registration phase} and it exploits the hints from the support samples to offer the co-occurrence prior knowledge. However, during the \textit{evaluation phase}, the new classifier is shared by all query images, hence the introduced prior might be biased towards the content of the limited support samples, causing inferior generalization ability to different query images. To alleviate this issue, we take a step further and propose the Dynamic Query Contextual Enrichment (DQCE) that adapts the new classifier to different contexts by dynamically incorporating essential semantic information mined from individual query samples. 

Specifically, we first let the original classifier to yield a temporary prediction $\vec{y}_{qry} \in \mathbb{R}^{h_q w_q\times N^b}$ on the query feature $\mathcal F(\vec{q}) \in \mathbb{R}^{h_q w_q\times d}$, where $h_q$, $w_q$, $N^b$ and $d$ are height, width, base class number and dimension number respectively. Then, the categorical representatives $\vec{p}^{b}_{qry} \in \mathbb{R}^{N^b \times d}$ of query sample $\vec{Q}$ are yielded as
\begin{equation}
    \footnotesize
	\label{eqn:query_proto}
	\vec{p}^{b}_{qry} = Softmax(\vec{y}_{qry}^{t}) \times \mathcal F(\vec{q}), 
\end{equation}
where the Softmax operation is performed on the second dimension of the transposed $\vec{y}_{cls}$ ($\vec{y}_{cls}^{t} \in \mathbb{R}^{N^b \times h_q w_q}$), so as to weigh the elements on the query feature map $\mathcal F(\vec{q})$.

Since the query label is not available, the dynamic prototypes $\vec{p}^{b}_{qry}$ yielded via query predictions are not as sound as $\vec{p}^{b}_{sup}$ and might introduce unnecessary noises. We still need a reliability measurement $\vec{\gamma}^i_{qry}$ for the estimated $\vec{p}^{b}_{qry}$.
Similar to Eq.~\eqref{eqn:capl_base_proto}, a weighing factor $\vec{\gamma}^i_{qry} = \mathcal G_{qry}(\vec{p}^{b, i}_{cls}, \vec{p}^{b, i}_{qry})$ is required to accomplish the dynamic context enrichment for each base class. Then, we can obtain the dynamically enriched prototypes as
\begin{equation}
    \footnotesize
	\label{eqn:capl_dynamic_proto}
	\vec{p}^{b, i}_{dyn} = \vec{\gamma}^i_{qry} * \vec{p}^{b, i}_{cls} + (1 - \vec{\gamma}^i_{qry}) * \vec{p}^{b, i}_{qry}
	, \quad i\in\{1,...,N^b\}.
\end{equation}
It is worth noting that, different from Eq.~\eqref{eqn:capl_base_proto} where only $n^b$ classes contained in the support set are enriched,  Eq.~\eqref{eqn:capl_dynamic_proto} considers all $N^b$ base classes since there is no prior knowledge of what classes are contained in the current query image. Besides, we empirically find the cosine similarity is better for $\mathcal G_{qry}$ and MLP is more suitable for $\mathcal G_{sup}$. Ablations on other alternatives are in the supplementary.

\begin{equation}
    \footnotesize
	\label{eqn:enhanced_classifier}
	\vec{p}^{b, i}_{capl} =  \vec{p}^{b, i} + \vec{p}^{b, i}_{dyn}
	, \quad i\in\{1,...,N^b\}.
\end{equation}
$\vec{P}^{b}_{capl}$ is then concatenated with the prototypes $\vec{P}^{n}$ of novel classes to form $\vec{P}^{all}_{capl} \in \mathbb{R}^{N^{b}+N^{n}, d}$, as to to predict all possible base and novel classes during the \textit{evaluation phase}.

\paragraph{Training.} 
Directly applying the weighted sum in Eq.~\eqref{eqn:capl_base_proto} and Eqs.~\eqref{eqn:capl_dynamic_proto}-\eqref{eqn:enhanced_classifier} is difficult because the averaged support/query features and the classifier's weights are in different feature spaces. To make it tractable, we accordingly modify the training scheme to let the feature extractor $\mathcal F$ learn to yield features that are compatible with $\vec{P}^{b}_{cls}$.

For getting prior knowledge from support samples, let $B$ denote the training batch size and $N^b$ denote the number of all base classes. We randomly select $\lfloor\frac{B}{2}\rfloor$ training samples as the `Fake Support' samples and the rest as `Fake Query' ones. Let $N^b_f$ denote the number of base classes contained in `Fake Support' samples. We randomly select $\lfloor\frac{N^b_f}{2}\rfloor$ as the `Fake Novel' classes $\vec{C}^{FN}$ and set the rest $N^b_f - \lfloor\frac{N^b_f}{2}\rfloor$ as the `Fake Context' classes $\vec{C}^{FC}$. The selected $\vec{C}^{FN}$ and $\vec{C}^{FC}$ classes are both base classes, but they mimic the behaviors of real novel and base classes contained in the support set respectively. The formation of the updated prototypes $\vec{p}^{b,i}$ ($i \in \{1, ..., N^b_f, ..., N^b\}$) during training can be written as
\begin{equation}
	\label{eqn:training_proto}
	\footnotesize
	\vec{p}^{b,i}=\left\{
	\begin{array}{lcl}
		\vec{\gamma}^i_{sup} * \vec{p}^{b, i}_{cls} + (1 - \vec{\gamma}^i_{sup}) * \vec{p}^{b, i}_{sup}				& & {c^i \in \vec{C}^{FC}}
		\\
		\vec{p}^{b,i}_{sup} 				& & {c^i \in \vec{C}^{FN}}
		\\
		\vec{p}^{b,i}_{cls} 				 & & \textrm{Otherwise}
	\end{array} \right.\\
\end{equation}
\vspace{-0.1cm}
\begin{equation}
	\label{eqn:training_new_proto}
	\footnotesize
	\vec{p}^{b, i}_{sup} = \frac
	{\sum_{j=1}^{\lfloor\frac{B}{2}\rfloor} \sum_{h,w} [\vec{m}_j^i \circ  \mathcal F(\vec{s}^i_j)]_{h,w}}   
	{\sum_{j=1}^{\lfloor\frac{B}{2}\rfloor} \sum_{h,w} [\vec{m}_j^i]_{h,w} }
	. \quad c^i \in \vec{C}^{FC} \cup \vec{C}^{FN}
\end{equation}
Specifically, `Fake Context' classes $\vec{C}^{FC}$ act as the base classes contained in support samples of novel classes during testing. The features of `Fake Context' classes update $\vec{P}^{b}_{cls}$ with $\vec{\gamma}_{sup}$ as shown in Eq. \eqref{eqn:training_proto}. Also, features of `Fake Novel' classes $\vec{C}^{FN}$ yield `Fake Novel' prototypes via Eq. \eqref{eqn:training_new_proto} that directly replace the learned weights of the corresponding base classes in $\vec{P}^{b}_{cls}$. The weights of the rest classes for this training batch are kept as the original ones in the classifier. 

On the other hand, in an attempt to make the classifier adapt to different contexts of individual query samples, the dynamic prototypes $\vec{p}^{b, i}_{dyn} (i\in\{1,...,N^b\})$ are generated by following Eqs.~\eqref{eqn:query_proto} and~\eqref{eqn:capl_dynamic_proto}. Finally, the context-aware classifier $\vec{P}^b_{capl}$ is formed as Eq.~\eqref{eqn:enhanced_classifier}, and the overall training objective is minimizing the standard cross-entropy loss calculated on the predictions made by $\vec{P}^b_{capl}$.

\section{Experiments}
\label{sec:experiments}
This section presents the experimental results, and the implementation details are shown in the supplementary file.

\subsection{Comparison with FS-Seg Models in GFS-Seg}
\label{sec:fs_seg_compare}
To show that models in the setting of FS-Seg are unable to perform well when predicting both base and novel classes, we evaluate four recently proposed exemplar FS-Seg frameworks (CANet~\cite{canet}, PFENet \cite{pfenet} SCL~\cite{scl} and PANet \cite{panet}) in the setting of GFS-Seg. 
The major difference between SCL, PFENet, CANet and PANet is about `Dist' shown in Figure \ref{fig:setting_compare} (a) relating to the method for processing query and support feature and making predictions -- SCL, PFENet are CANet are relation-based models and PANet is a cosine-based model. Specifically, CANet uses convolutions as a relation module \cite{relationnet} to process the concatenated query and support prototypes to yield prediction on the query image, but PANet generates results by measuring the cosine similarity between every query feature and prototypes got from support samples. Though FS-Seg only requires evaluation of novel classes, if the prototypes (averaged features) of base classes are available, the FS-Seg models should also be able to identify the regions of base classes in query images. 
It is worth noting that some recently proposed advanced frameworks, \eg, RePRI~\cite{repri} and HSNet~\cite{hsnet}, are not practical to be applied to the setting of GFS-Seg because their inference processes require independent reasoning with the full support feature maps, instead of the prototypical feature vectors adopted by above methods (the prior mask cannot be used in PFENet). It means that under an $n$-way $k$-shot setting, for each query image, they need to perform the entire feature reasoning for additional $nk$ times on the full support feature maps, causing low efficiency or even OOM (Out of Memory) issues.

We use the publicly available codes 
	and follow the default training configurations. We modify the inference code to feed all prototypes (base and novel) for each query image. The base prototypes are formed by averaging the features belonging to base classes in all training samples. As shown in Table \ref{tab:compare_fsseg_new},  CANet, SCL, PFENet, and PANet do not perform well compared to the models implemented with CAPL. `PANet + CAPL' differs from PANet only in the training and classifier formation strategies.

We note that the results of novel mIoU in Table \ref{tab:compare_fsseg_new} are under the GFS-Seg setting and thus they are lower than that reported in the papers of these FS-Seg models.
This discrepancy is caused by different settings. In GFS-Seg, models are required to identify all classes in a given testing image, including both base and novel classes, while in FS-Seg, models only need to find the pixels belonging to one specific novel class with support samples that provide the prior knowledge of what the target class is. Therefore, it is much harder to identify the novel classes under the interference of the base classes in GFS-Seg.

More specifically, the main reason that FS-Seg models fall short is that the episodic training/testing schemes of FS-Seg only focus on adapting models to be discriminative between background and foreground, where the decision boundary for each episode only lies between one target class and the background in each query sample. Also, FS-Seg requires query images to contain the classes provided by support samples, while GFS-Seg distinguishes between not only multiple novel classes but also all possible base classes simultaneously without the prior knowledge of what classes are contained in query samples. Moreover, to maintain the high generalization ability on unseen novel classes in FS-Seg, both CANet, SCL and PFENet fix their backbones during training, causing poor adaptation in the complex scenario of GFS-Seg that requires multi-class labeling.

\begin{table}[!t]
	\scriptsize
	\vspace{-0.2cm}
	\centering
	\tabcolsep=0.175cm
	{
		\begin{tabular}{ 
				l
				c  c  c 
				c  c  c }
			\toprule
			&\multicolumn{3}{c}{1-shot} 
			&\multicolumn{3}{c}{5-shot}  
			\\  		
			\specialrule{0em}{0pt}{1pt}			
			\cline{2-7}
			\specialrule{0em}{1pt}{0pt}			
			
			Methods
			& Base  & Novel  & Total 
			& Base  & Novel  & Total \\
			
			\specialrule{0em}{0pt}{1pt}
			\hline
			\specialrule{0em}{1pt}{0pt}
			
			CANet~\cite{canet}
			& 8.73 & 2.42 & 7.23
			& 9.05 & 1.52 & 7.26 \\        
			
			PFENet~\cite{pfenet} 
			& 8.32 & 2.67 & 6.97 
			& 8.83 & 1.89 & 7.18 \\     
			
			SCL~\cite{canet} 
			& 8.88 & 2.44 & 7.35
			& 9.11 & 1.83 & 7.38 \\					
			
			PANet~\cite{panet}  
			& 31.88 & 11.25 & 26.97
			& 32.95 & 15.25 & 28.74 \\
			
			PANet + CAPL 
			& 63.06 & 14.96 & 51.60
			& 63.81 & 19.66 & 53.30 \\ 
			
			DeepLab-V3 + CAPL
			& 65.71 & 15.05 & 53.77
			& 67.01 & 23.26 & 56.59 \\ 
			
			PSPNet + CAPL 
			& 65.48 & 18.85 & 54.38
			& 66.14 & 22.41 & 55.72  \\ 			
			
			\bottomrule      			                             
		\end{tabular}
	}
	\caption{Comparisons with FS-Seg models in GFS-Seg where the base and novel classes are required to be simultaneously identified. All models are based on ResNet-50. \textbf{Base}: mIoU results of all base classes. \textbf{Novel}: mIoU results of all novel classes. \textbf{Total}: mIoU results of all (base + novel) classes.}	
	\label{tab:compare_fsseg_new}   	
	\vspace{-0.3cm}
\end{table}

\begin{table}[!t]
	\scriptsize
	\centering
	\tabcolsep=0.175cm
	{
		\begin{tabular}{ 
				l
				c  
				c 
				c  c  c 
				c  c  c }
			\toprule
			& 
			&
			&\multicolumn{3}{c}{1-shot} 
			&\multicolumn{3}{c}{5-shot}  
			\\  		
			\specialrule{0em}{0pt}{1pt}			
			\cline{4-9}
			\specialrule{0em}{1pt}{0pt}			
			
			Methods
			& MLP
			& Cos
			& Base  & Novel  & Total 
			& Base  & Novel  & Total \\
			
            \specialrule{0em}{0pt}{1pt}
            \hline
            \specialrule{0em}{1pt}{0pt}
            \multicolumn{9}{c}{Pascal-5$^i$}  \\
            \specialrule{0em}{0pt}{1pt}
            \hline
            \specialrule{0em}{1pt}{0pt} 
			
			Baseline & N/A     
			& N/A    
			& 60.47 & 14.55 & 49.54
			& 61.88 & 16.68 & 51.12 \\        
			DQCE  & \Checkmark 
			& -   
			& 63.25 & 15.42 & 51.82 
			& 64.12 & 20.37 & 53.70 \\  	
			DQCE  & -    
			& \Checkmark 
			& 64.16 & 15.39 & 52.55 
			& 65.26 & 21.32 & 54.80 \\  
			
			DQCE-Sw  & -    
			& \Checkmark 
			& 58.28	& 15.96 & 48.20
			& 60.68 & 20.66 & 51.15 \\ 			
			
			SCE  & \Checkmark 
			& -   
			& 62.17 & 17.88 & 51.63
			& 63.62 & 20.50 & 53.35 \\  			
			
			SCE  & -    
			& \Checkmark 
			& 59.87 & 16.80 & 49.62 
			& 61.60 & 19.92 & 51.68 \\  
			
			SCE-Sw  & -    
			& \Checkmark 
			& 60.92 & 16.39 & 50.32
			& 62.75 & 21.18 & 52.83 \\

    		CAPL  & N/A 
			& N/A   
			& \textbf{65.48} & \textbf{18.85} & 	\textbf{54.38} 
			& \textbf{66.14} & \textbf{22.41} & 	\textbf{55.72}    \\  					

            \specialrule{0em}{0pt}{1pt}
            \hline
            \specialrule{0em}{1pt}{0pt}
            \multicolumn{9}{c}{COCO-20$^i$}  \\
            \specialrule{0em}{0pt}{1pt}
            \hline
            \specialrule{0em}{1pt}{0pt} 
            
			Baseline & N/A     
			& N/A    
			& 36.68 & 5.84 &  29.06 
			& 36.91 & 7.26 &  29.59 \\          
			
			CAPL  & N/A 
			& N/A   
			& \textbf{44.61} & \textbf{7.05} & \textbf{35.46}
			& \textbf{45.24} & \textbf{11.05} & \textbf{36.80} \\  				

			\bottomrule      			                             
		\end{tabular}
	}
	\vspace{-0.2cm}	
	\caption{Comparison of contextual enrichment strategies. `MLP' and 'Cos' mean two-layers MLPs and cosine similarity are used for yielding weighing factors $\vec{\gamma}_{qry}$ and $\vec{\gamma}_{sup}$. SCE mines co-occurrence cues from support data (Eqs.~\eqref{eqn:capl_base_new_proto}-\eqref{eqn:capl_base_proto}, and 'DQCE' extracts temporary query context  (Eqs.~\eqref{eqn:query_proto}-\eqref{eqn:capl_dynamic_proto}). `CAPL' combines `SCE (MLP)' and 'DQCE (Cos)' as Eq.~\eqref{eqn:enhanced_classifier}. }  
	\label{tab:ablation_context}  	
	\vspace{-0.1cm}
\end{table}

\begin{table}[!t]
	\scriptsize
	\vspace{-0.2cm}
	\centering
	\tabcolsep=0.175cm
	{
		\begin{tabular}{ 
				l
				c  
				c 
				c  c  c 
				c  c  c }
			\toprule
			& 
			&
			&\multicolumn{3}{c}{1-shot} 
			&\multicolumn{3}{c}{5-shot}  
			\\  		
			\specialrule{0em}{0pt}{1pt}			
			\cline{4-9}
			\specialrule{0em}{1pt}{0pt}			
			
			Methods
			& Train
			& Test
			& Base  & Novel  & Total 
			& Base  & Novel  & Total \\
			
			\specialrule{0em}{0pt}{1pt}
			\hline
			\specialrule{0em}{1pt}{0pt}
			
			Baseline & N/A     
			& N/A    
			& 60.47 & 14.55 & 49.54  
			& 61.88 & 16.68 & 51.12 \\            
			
			Baseline+  & N/A 
			& N/A   
			& 60.60 & 16.53 & 50.10
			& 62.28 & 19.39 & 52.07 \\  			
			
			CAPL-Tr  & \Checkmark 
			& -   
			& 59.73 & 17.40 & 49.65 
			& 61.34	& 21.72 & 51.91 \\  				
			
			CAPL-Te  & -    
			& \Checkmark 
			& 60.89 & 7.00 & 48.06
			& 61.13 & 10.90 & 49.17 \\  
			
			CAPL  & \Checkmark 
			& \Checkmark   
			& \textbf{65.48} & \textbf{18.85} & 	\textbf{54.38} 
			& \textbf{66.14} & \textbf{22.41} & 	\textbf{55.72}    \\

			\bottomrule      			                             
		\end{tabular}
	}
	\caption{Comparison of training \& testing strategies of CAPL. `Baseline+' merely replaces fake novel prototypes during training. `CAPL-Tr' adopts CAPL's training strategy while keeping the novel class registration and evaluation phases the same as the baseline.
	`CAPL-Te' performs CAPL during the novel class registration and evaluation phases, but its training scheme is not altered. }	
	\label{tab:ablation_training}    	
	\vspace{-0.35cm}
\end{table}

\subsection{Ablation Study}
\label{sec:ablation_study}
In this section, we investigate the components of CAPL with PSPNet on Pascal-$5^i$ in GFS-Seg where the base and novel classes are required to be simultaneously identified. 
The baselines for the following ablation study are based on PSPNet~\cite{pspnet} with ResNet-50~\cite{resnet}.

\paragraph{Design options for components of CAPL. } 
The effects of the support contextual enrichment method (SCE) as well as its dynamic counterpart (DQCE) are investigated in Table~\ref{tab:ablation_context}.
Concretely, SCE enriches the classifier with the essential co-occurrence relations between novel and base classes (Eqs.~\eqref{eqn:capl_base_new_proto}-\eqref{eqn:capl_base_proto}), while DQCE further adapts the enrichment process to the content of individual query images (Eqs.~\eqref{eqn:query_proto}-\eqref{eqn:enhanced_classifier}). Thus, $\mathcal G_{sup}$ and $\mathcal G_{qry}$ are used by SCE and DQCE to yield $\vec{\gamma}_{sup}$ and $\vec{\gamma}_{qry}$ respectively. It can be observed in Table~\ref{tab:ablation_context} that both SCE and DQCE are most conducive to the baseline when $\mathcal G_{qry}$=`Cos' and $\mathcal G_{sup}$=`MLP'.
Cosine similarity works better for DQCE because it serves as the reliability estimator to weigh the prototypes extracted from the query features as they sometimes introduce irrelevant information. However, MLP is more likely to give high confidence to the original classifier to avoid the risk brought by Eq.~\eqref{eqn:query_proto}, yielding suboptimal results. Also, switching the weighing factors for $\vec{p}^{b, i}_{cls}$ and $\vec{p}^{b, i}_{qry}$ (DQCE-Sw) undoubtedly worsens the performance because the new prototype will be dominated by untrustworthy $\vec{p}^{b, i}_{qry}$.

Nonetheless, cosine similarity is less effective than MLP for SCE, and `Cos' achieves comparable results to the baseline. This could be caused by the fact that $\vec{p}^{b, i}_{sup}$ is produced by ground-truth masks, hence $\vec{p}^{b, i}_{sup}$ is a reliable categorical representation, leading to a high similarity value with $\vec{p}^{b, i}_{cls}$ that directly suppresses the context brought by $\vec{p}^{b, i}_{sup}$ in Eq.~\eqref{eqn:capl_base_proto}. If we switch the weighing factors for $\vec{p}^{b, i}_{cls}$ and $\vec{p}^{b, i}_{sup}$ in Eq.~\eqref{eqn:capl_base_proto} (`SCE-Sw' in Table~\ref{tab:ablation_context}), the results will more rely on the content of $\vec{p}^{b, i}_{sup}$, losing the generalization power of $\vec{p}^{b, i}_{cls}$. Differently, we observe that values of $\vec{\gamma}_{sup}$ yielded by MLP are within a certain range (0.4-0.7) without corrupting to 0 or 1, but they are still generally proportional to the discrepancy between $\vec{p}^{b,i}_{cls}$ and $\vec{p}^{b,i}_{sup}$ of base class $i$. Thus, SCE (MLP) outweighs SCE-Sw (Cos) by producing moderate data-conditioned values to adequately balance the novel contextual and original information.

\paragraph{Training and contextual enrichment strategies. } 
Since the training of CAPL picks `Fake Novel' and `Fake Context' samples to make the averaged features compatible with learned weights for building the updated classifier, we modify the training scheme of the baseline (denoted as `Baseline+') accordingly for a fair comparison. Because the baseline only replaces the novel prototypes during the evaluation, only `Fake Novel' is sampled and used to replace the base prototypes during training Baseline+. Therefore, Baseline+ is analogous to the methods in few-shot classification~\cite{imprint_cvpr18,dynamic_noforget}. Besides, `CAPL-Tr' denotes that CAPL is only performed during the training phase, keeping the novel class registration and evaluation phases intact as the baseline.  `CAPL-Te' represents that CAPL is only performed during the \textit{novel class registration} and \textit{evaluation phases} and there is no modification to the baseline training. 
Since $\vec{\gamma}_{sup}$ is yielded by a trainable two-layers MLP, we set $\vec{\gamma}_{sup}$ to the mean converged values of CAPL for CAPL-Te.

The results in Table \ref{tab:ablation_training} show that the training and contextual enrichment strategies of CAPL complement each other -- both are indispensable. CAPL-Tr brings minor improvement to the baseline with only the training alignment is implemented, and CAPL-Te proves that, without the proposed training strategy, the contextual enrichment method of CAPL produces sub-optimal results due to misalignment between the original classifier and the features.

\begin{table}
	\scriptsize
	\centering
	\tabcolsep=0.06cm
	{
		\begin{tabular}{ 
				l
				c
				c 
				c
				c 
				c 
				c 
				c 
				c 
				c 
				c 
				c
				c 
			}
			\toprule     
			& 
			&
			& 
			&			
			& 
			&\multicolumn{3}{c}{Pascal-5$^i$} 
			& 
			&\multicolumn{3}{c}{COCO-20$^i$}  
			\\  
			\textit{Methods}
			& 
			& Venue
			&
			& Backbone
			&
			& 1-Shot
			&
			& 5-Shot
			&
			& 1-Shot
			&
			& 5-Shot 		\\
			
			\specialrule{0em}{0pt}{1pt}
			\hline
			\specialrule{0em}{1pt}{0pt}

			\specialrule{0em}{0pt}{1pt}
			\hline
			\specialrule{0em}{1pt}{0pt}

			PANet~\cite{panet}
			& 
			& ICCV-19
			&     
			& Res-50
			&
			& 48.1
			& 
			& 55.7
			&
			& 20.9           
			&
			& 29.7 		\\

			PFENet \cite{pfenet}
			& 
			& TPAMI-20
			&     
			& Res-50
			&
			& 60.8
			& 
			& 61.9
			&
			& 32.1          
			&
			& 37.5	\\   	
			
		    ASGNet \cite{asgnet}
			& 
			& CVPR-21
			&     
			& Res-50
			&
			& 59.3
			& 
			& 63.9
			&
			& 34.5   
			&
			& 42.5 	\\  			
			
		    SCL \cite{scl}
			& 
			& CVPR-21
			&     
			& Res-50
			&
			& 61.8
			& 
			& 62.9
			&
			& -          
			&
			& - 	\\         		
		
		    SAGNN \cite{sagnn}
			& 
			& CVPR-21
			&     
			& Res-50
			&
			& 62.1
			& 
			& 62.8
			&
			& -          
			&
			& - 	\\    		
			
		    RePri \cite{repri}
			& 
			& CVPR-21
			&     
			& Res-50
			&
			& 59.1
			& 
			& 66.8
			&
			& 34.0          
			&
			& 42.1 	\\      	
			
		    CWT~\cite{cwt}
			& 
			& ICCV-21
			&     
			& Res-50
			&
			& 56.4
			& 
			& 63.7
			&
			& 32.9      
			&
			& 41.3 	\\ 	
			
		    MMNet~\cite{mmnet}
			& 
			& ICCV-21
			&     
			& Res-50
			&
			& 61.8
			& 
			& 63.4
			&
			& 37.5  
			&
			& 38.2 	\\ 				
			
		    CMN~\cite{cmn}
			& 
			& ICCV-21
			&     
			& Res-50
			&
			& \underline{62.8}
			& 
			& 63.7
			&
			& 39.3     
			&
			& 43.1 	\\ 				
			
			Mining~\cite{mining}
			& 
			& ICCV-21
			&     
			& Res-50
			&
			& 62.1
			& 
			& 66.1
			&
			& 33.9         
			&
			& 40.6 	\\ 			
			
		    HSNet \cite{hsnet}
			& 
			& ICCV-21
			&     
			& Res-50
			&
			& \textbf{64.0}
			& 
			& \textbf{69.5}
			&
			& \underline{39.2}
			&
			& 46.9 	\\

			\textbf{CAPL (PANet)}
			& 
			& 
			&     
			& Res-50
			&
			& 60.6
			& 
			& 66.1
			&
			& 38.0      
			&
			& \underline{47.3} 	\\       
			
			\textbf{CAPL (PFENet)}
			& 
			& 
			&     
			& Res-50
			&
			& 62.2  
			& 
			& \underline{67.1}
			&
			& \textbf{39.8}           
			&
			& \textbf{48.3} 		\\    			
			
			\specialrule{0em}{0pt}{1pt}
			\hline
			\specialrule{0em}{1pt}{0pt}

			PFENet \cite{pfenet}
			& 
			& TPAMI-20
			&     
			& Res-101
			&
			& 60.1
			& 
			& 61.4
			&
			& 32.4           
			&
			& 37.4 	\\      	  

		    SAGNN \cite{sagnn}
			& 
			& CVPR-21
			&     
			& Res-101
			&
			& -
			& 
			& -
			&
			& 37.2         
			&
			& 42.7 	\\   

		    ASGNet \cite{asgnet}
			& 
			& CVPR-21
			&     
			& Res-101
			&
			& 59.3
			& 
			& 64.4
			&
			& -  
			&
			& - 	\\  

		    CWT~\cite{cwt}
			& 
			& ICCV-21
			&     
			& Res-101
			&
			& 58.0
			& 
			& 64.7
			&
			& 32.4        
			&
			& 42.0 	\\ 		
			
			Mining~\cite{mining}
			& 
			& ICCV-21
			&     
			& Res-101
			&
			& 62.6
			& 
			& 68.8
			&
			& 36.4         
			&
			& 44.4 	\\ 					
			
		    HSNet \cite{hsnet}
			& 
			& ICCV-21
			&     
			& Res-101
			&
			& \textbf{66.2}
			& 
			& \textbf{70.4}
			&
			& \underline{41.2}
			&
			& \underline{49.5} 	\\

			\textbf{CAPL (PFENet)}
			& 
			& 
			&     
			& Res-101
			&
			& \underline{63.6}
			& 
			& \underline{68.9} 
			&
			& \textbf{42.8}           
			&
			& \textbf{50.4} \\                
			\bottomrule                    
		\end{tabular}
	}
	\caption{Class mIoU results on Pascal-5$^i$ and COCO-20$^i$ in FS-Seg where only the novel classes are required to be identified.
	}          	
	\label{tab:compare_panet} 	
	\vspace{-0.35cm}
\end{table}

\subsection{Apply CAPL to FS-Seg}
FS-Seg is an extreme case of GFS-Seg. 
To validate the proposed CAPL in the setting of FS-Seg, in Table \ref{tab:compare_panet}, we incorporate CAPL to PANet and PFENet, and CAPL has achieved significant improvement to the baselines. Specifically, CAPL only alters the prototype construction process of PANet by making it dynamically adapt to different query and support pairs, following the method introduced in Section~\ref{sec:method}. Besides, we also incorporate CAPL to PFENet whose decoder processes the concatenation of the prior mask and middle-level features to make predictions. However, the prior mask is enhanced by CAPL. Since there is no significant structural change on both PANet and PFENet, the improved models have nearly the same inference speed as the original ones.
More implementation details regarding this section are shown in the supplementary.

\subsection{Visual Examples}
\begin{figure}[!t]
    \scriptsize
	\centering
	\begin{minipage}  {0.185\linewidth}
		\centering
		\includegraphics [width=1\linewidth,height=0.8\linewidth] 
		{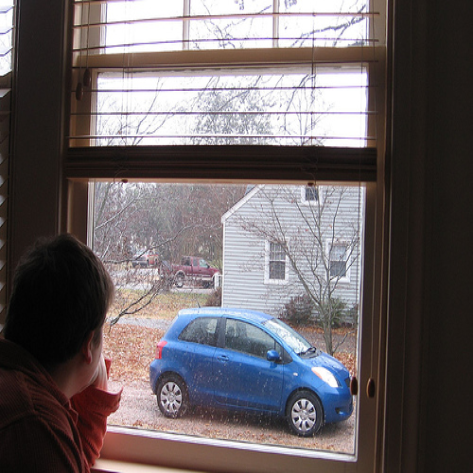}
	\end{minipage}      
	\begin{minipage}  {0.185\linewidth}
		\centering
		\includegraphics [width=1\linewidth,height=0.8\linewidth] 
		{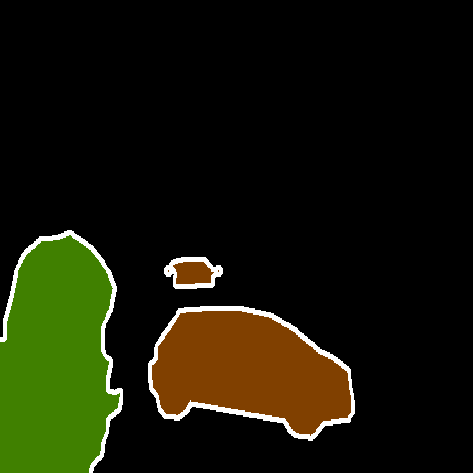}
	\end{minipage}      
	\begin{minipage}  {0.185\linewidth}
		\centering
		\includegraphics [width=1\linewidth,height=0.8\linewidth] 
		{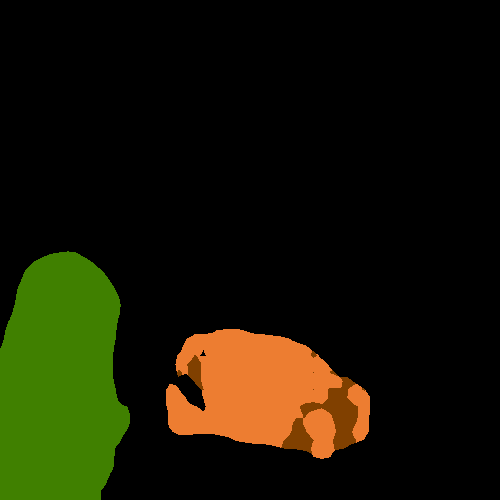}
	\end{minipage}     	
	\begin{minipage}  {0.185\linewidth}
		\centering
		\includegraphics [width=1\linewidth,height=0.8\linewidth] 
		{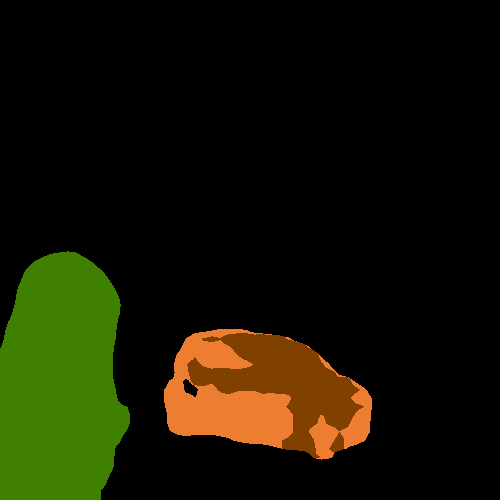}
	\end{minipage}     
	\begin{minipage}  {0.185\linewidth}
		\centering
		\includegraphics [width=1\linewidth,height=0.8\linewidth] 
		{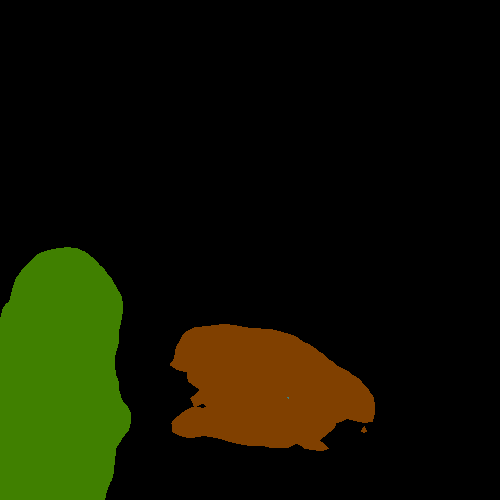}
	\end{minipage}     	

	\begin{minipage}  {0.185\linewidth}
		\centering
		\includegraphics [width=1\linewidth,height=0.8\linewidth] 
		{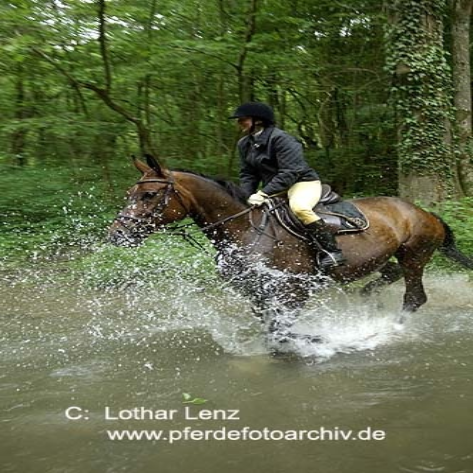}
	\end{minipage}      
	\begin{minipage}  {0.185\linewidth}
		\centering
		\includegraphics [width=1\linewidth,height=0.8\linewidth] 
		{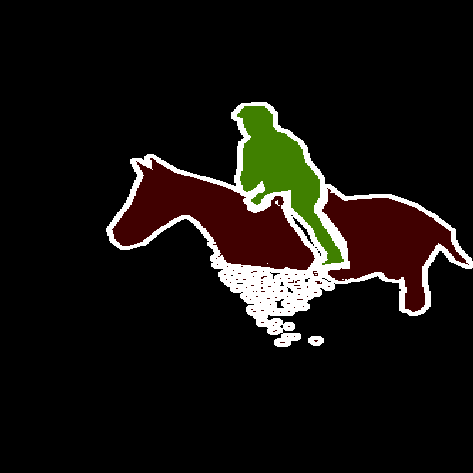}
	\end{minipage}      
	\begin{minipage}  {0.185\linewidth}
		\centering
		\includegraphics [width=1\linewidth,height=0.8\linewidth] 
		{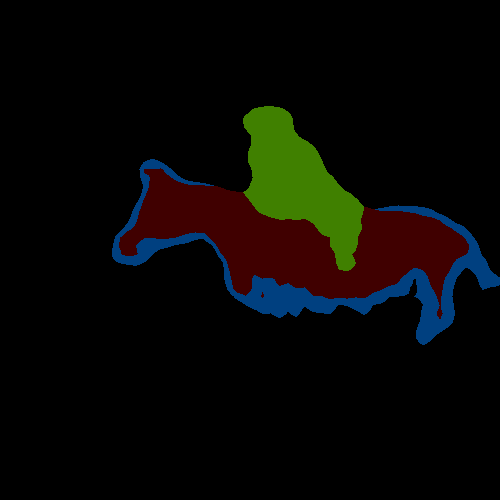}
	\end{minipage}     
	\begin{minipage}  {0.185\linewidth}
		\centering
		\includegraphics [width=1\linewidth,height=0.8\linewidth] 
		{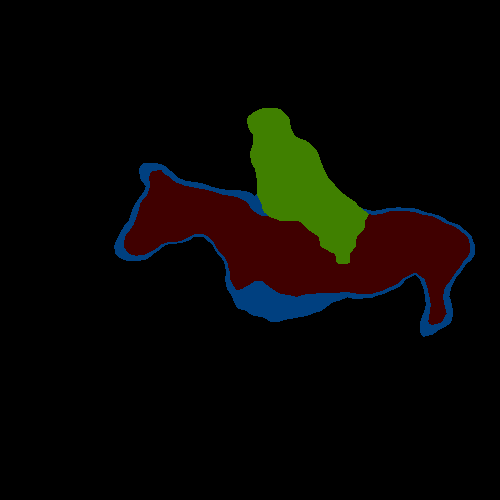}
	\end{minipage}     	
	\begin{minipage}  {0.185\linewidth}
		\centering
		\includegraphics [width=1\linewidth,height=0.8\linewidth] 
		{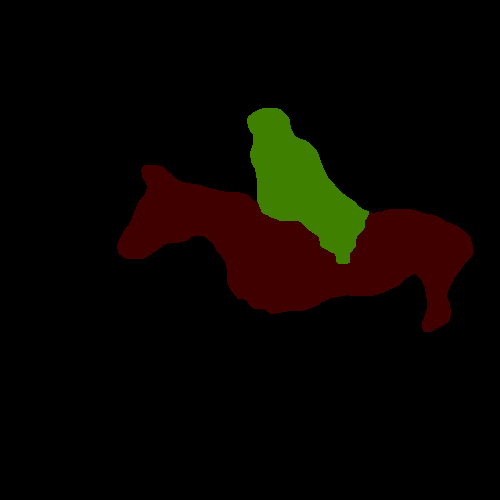}
	\end{minipage}     	

	\begin{minipage}  {0.185\linewidth}
		\centering
		\includegraphics [width=1\linewidth,height=0.8\linewidth] 
		{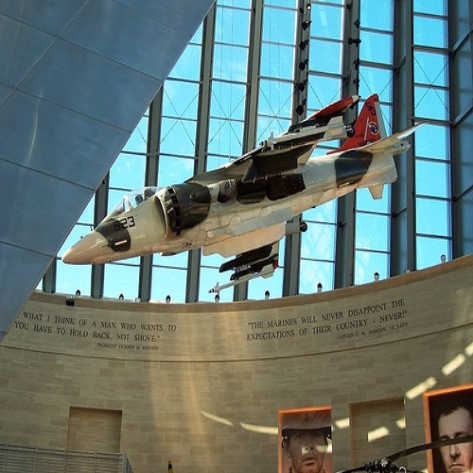}
	\end{minipage}      
	\begin{minipage}  {0.185\linewidth}
		\centering
		\includegraphics [width=1\linewidth,height=0.8\linewidth] 
		{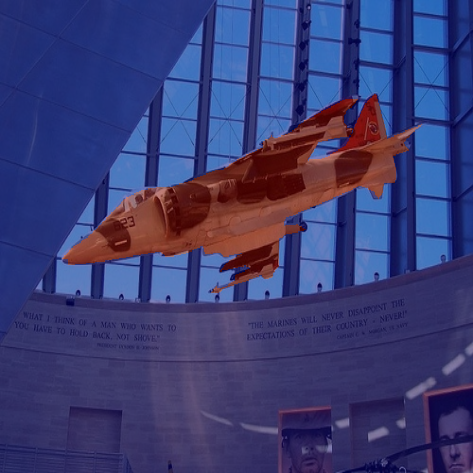}
	\end{minipage}      
	\begin{minipage}  {0.185\linewidth}
		\centering
		\includegraphics [width=1\linewidth,height=0.8\linewidth] 
		{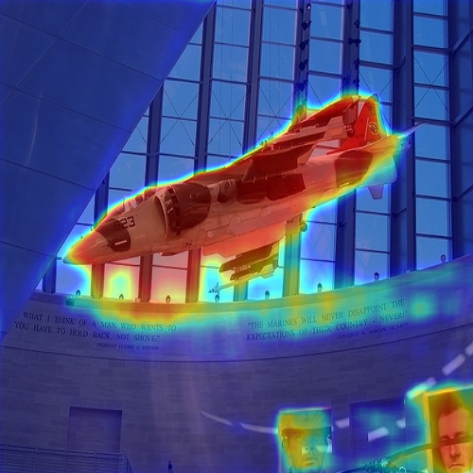}
	\end{minipage}     
	\begin{minipage}  {0.185\linewidth}
		\centering
		\includegraphics [width=1\linewidth,height=0.8\linewidth] 
		{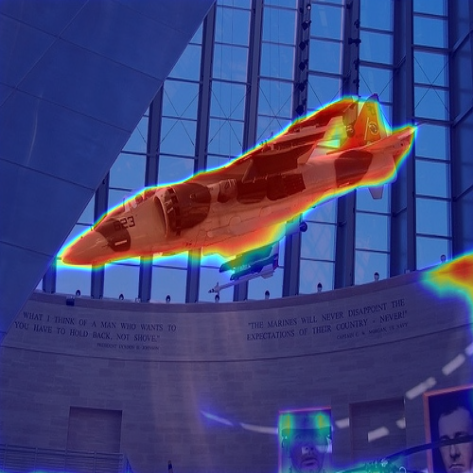}
	\end{minipage}     	
	\begin{minipage}  {0.185\linewidth}
		\centering
		\includegraphics [width=1\linewidth,height=0.8\linewidth] 
		{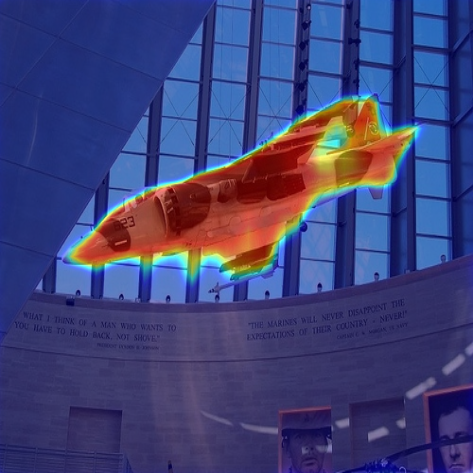}
	\end{minipage}     	

	\begin{minipage}  {0.185\linewidth}
		\centering
		\includegraphics [width=1\linewidth,height=0.8\linewidth] 
		{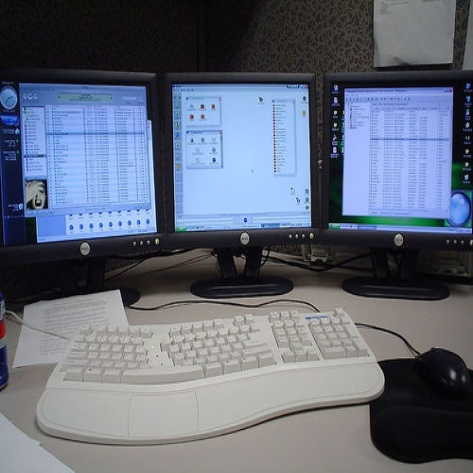}
		{\footnotesize{(a)}}
	\end{minipage}      
	\begin{minipage}  {0.185\linewidth}
		\centering
		\includegraphics [width=1\linewidth,height=0.8\linewidth] 
		{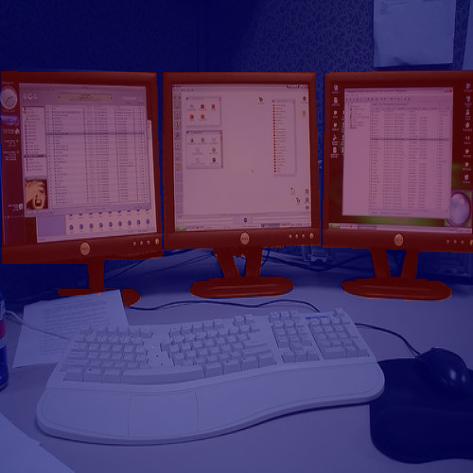}
		{\footnotesize{(b)}}
	\end{minipage}      
	\begin{minipage}  {0.185\linewidth}
		\centering
		\includegraphics [width=1\linewidth,height=0.8\linewidth] 
		{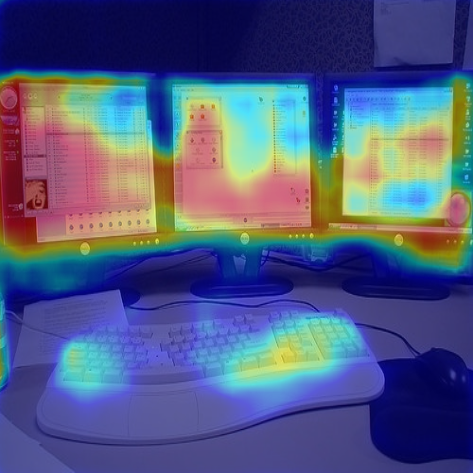}
		{\footnotesize{(c)}}
	\end{minipage}     
	\begin{minipage}  {0.185\linewidth}
		\centering
		\includegraphics [width=1\linewidth,height=0.8\linewidth] 
		{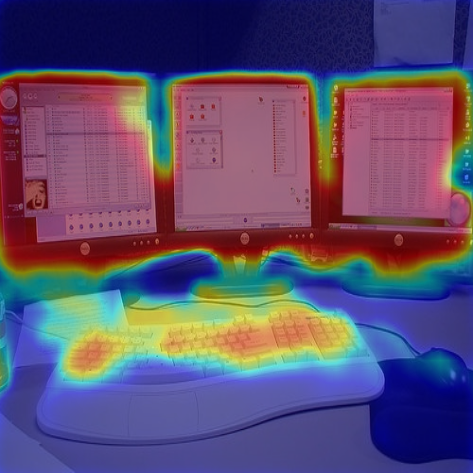}
		{\footnotesize{(d)}}
	\end{minipage}     	
	\begin{minipage}  {0.185\linewidth}
		\centering
		\includegraphics [width=1\linewidth,height=0.8\linewidth] 
		{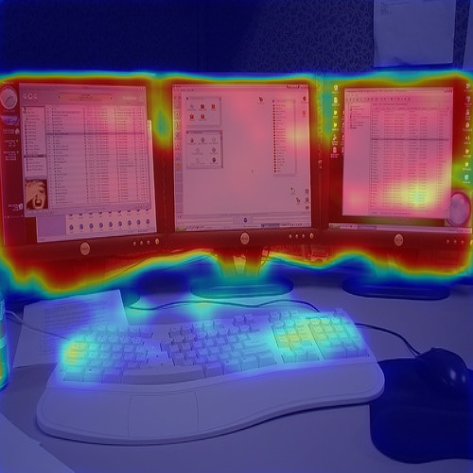}		
		{\footnotesize{(e)}}
	\end{minipage}    	
	
	\vspace{-0.3cm}
	\caption{Visualizations of GFS-Seg (top) and FS-Seg (bottom). (a): Input; (b): GT; (c): Baseline; (d): SCE; (e): SCE+DQCE. }
	\vspace{-0.3cm}
	\label{fig:visual_compare_fsseg}
\end{figure}

Visual comparison is presented in Fig~\ref{fig:visual_compare_fsseg} where SCE and DQCE refine the baseline predictions. More visualizations are shown in the supplementary file. 

\section{Concluding Remarks}
\paragraph{Summary.}
We have presented the new benchmark of Generalized Few-Shot Semantic Segmentation (GFS-Seg) with a novel solution -- Context-Aware Prototype Learning (CAPL). 
Different from the classic Few-Shot Segmentation (FS-Seg), GFS-Seg aims at identifying both base and novel classes that FS-Seg models fall short. Our proposed CAPL achieves significant performance improvement by dynamically enriching the context information with adapted features. 
CAPL has no structural constraints on the base model and thus it can be easily applied to normal semantic segmentation frameworks, and it generalizes well to FS-Seg.

\vspace{-0.15cm}
\paragraph{Limitation \& Discussion with HSNet.}
CAPL dynamically leverages the contextual hints in both GFS-Seg and FS-Seg, while it does not introduce new designs for dense spatial reasoning between query and support features in FS-Seg. Thus, though new SOTA performance has been achieved on the challenging COCO-20$^i$ (20 novel classes) where the semantic cues are better exploited, CAPL does not outperform another advanced method that adopts the hyper-correlations between query and support features, \ie, HSNet~\cite{hsnet}, on Pascal-5$^i$ (5 novel classes) in FS-Seg.

{\small
	\bibliographystyle{ieee_fullname}
	\bibliography{egbib}
}

\end{document}